\documentclass{article}
\usepackage[utf8]{inputenc}

\usepackage{listings}
\usepackage{xcolor}

\definecolor{codegreen}{rgb}{0,0.6,0}
\definecolor{codegray}{rgb}{0.5,0.5,0.5}
\definecolor{codepurple}{rgb}{0.58,0,0.82}
\definecolor{backcolour}{rgb}{0.95,0.95,0.92}

\lstdefinestyle{mystyle}{
    backgroundcolor=\color{backcolour},   
    commentstyle=\color{codegreen},
    keywordstyle=\color{magenta},
    numberstyle=\tiny\color{codegray},
    stringstyle=\color{codepurple},
    basicstyle=\ttfamily\footnotesize,
    breakatwhitespace=false,         
    breaklines=true,                 
    captionpos=b,                    
    keepspaces=true,                 
    numbers=left,                    
    numbersep=3pt,                  
    showspaces=false,                
    showstringspaces=false,
    showtabs=false,                  
    tabsize=2
}
\usepackage{listings}

\usepackage[utf8]{inputenc}
\usepackage{listings,xcolor}
\usepackage{hyperref}
\usepackage{graphicx}
\usepackage{float}
\usepackage[T1]{fontenc}
\usepackage{xcolor}
\usepackage[scaled=0.69]{DejaVuSansMono}
\definecolor{commentgreen}{RGB}{2,112,10}
\definecolor{eminence}{RGB}{108,48,130}
\definecolor{weborange}{RGB}{255,165,0}
\definecolor{frenchplum}{RGB}{129,20,83}

\lstdefinelanguage{elixir}{
    morekeywords={case,catch,def,do,else,false,%
        use,alias,receive,timeout,defmacro,defp,%
        for,if,import,defmodule,defprotocol,%
        nil,defmacrop,defoverridable,defimpl,%
        super,fn,raise,true,try,end,with,%
        unless},
    otherkeywords={<-,->, |>, \%\{, \}, \{, \, \(, \)},
    sensitive=true,
    morecomment=[l]{\#},
    morecomment=[n]{/*}{*/},
    morecomment=[s][\color{purple}]{:}{\ },
    morestring=[s][\color{orange}]{Axon},
    commentstyle=\color{commentgreen},
    keywordstyle=\color{eminence},
    stringstyle=\color{yellow},
	basicstyle=\ttfamily,
	breaklines,
	showstringspaces=false
}
\title{Comparing neural network training performance between Elixir and Python}
\author{Lucas C. Tavano\footnote{E-mail: \href{mailto:lucas.c.tavano@gmail.com}{lucas.c.tavano@gmail.com}.}, Lucas K. Amin\footnote{E-mail: \href{mailto:lucas_amin_2@hotmail.com}{lucas\_amin\_2@hotmail.com}.} \, and Adolfo Gustavo Serra-Seca-Neto\footnote{E-mail: \href{mailto:adolfo@utfpr.edu.br}{adolfo@utfpr.edu.br}.}}
\date{October 2022}

\begin{document}

\maketitle

\begin{abstract}

With a wide range of libraries focused on the machine learning market, such as TensorFlow, NumPy, Pandas, Keras, and others, Python has made a name for itself as one of the main programming languages. In February 2021, José Valim and Sean Moriarity published the first version of the Numerical Elixir (Nx) library, a library for tensor operations written in Elixir. Nx aims to allow the language be a good choice for GPU-intensive operations. This work aims to compare the results of Python and Elixir on training convolutional neural networks (CNN) using MNIST and CIFAR-10 datasets, concluding that Python achieved overall better results, and that Elixir is already a viable alternative.




\end{abstract}

\section{Introduction}

Programming languages have optimized their data structures and libraries for processing n-dimensional tensors seeking performance and simplicity. Currently, the market for massively parallel data processing, such as machine learning, is led by the Python language, followed by \textit{R}, \textit{JavaScript}, and \textit{C++} \cite{allamanis2018survey}.

In this scenario, the Elixir language ecosystem has received new tools such as the Nx library, a new compiler to perform operations in GPU, the ability to work with float16 variables, among other improvements. These new features enables the Elixir language to participate in the machine learning market. This work aims to compare the programming languages Elixir and Python in the context of processing large-volume tensors, comparing execution time, process parallelization, and machine resources use like Central Processing Unit (CPU), Random Access Memory (RAM), Graphics Processing Unit (GPU), and Video RAM (VRAM) in similar algorithms.

\section{Conceptual and Empirical Foundations}
\label{chap:fundamentacao}

This chapter presents theoretical background on tensors, neural networks, datasets, and the Elixir Nx and the Python NumPy libraries. It also discusses the difference between functional and object-oriented paradigms in the context of this project and how this can impact the comparison between Elixir and Python results.

\subsection{Datasets}
\label{sec:funDataset}

A dataset is a large amount of data related to the same context \cite{IntroDatasets}. In this project, we will work with two datasets: MNIST and CIFAR-10, which will be used to train CNN's capable of classifying objects in images.

\subsubsection{What is the MNIST dataset}
\label{sec:funMnist}

MNIST is a labeled set of images of handwritten digits containing 70 thousand entries; 60 thousand represent the training set, and 10 thousand represent the test set. Each image in this collection has a resolution of 28x28 pixels; thus, it is represented in code by a 28 by 28 matrix. Each value contained in this matrix ranges from 0 to 255, 0 represents black, and 255 is white.

\begin{figure}[H]
    \centering
    \caption{Example of MNIST set images plotted graphically with the \textit{Matplotlib} tool}
    \includegraphics[width=\linewidth]{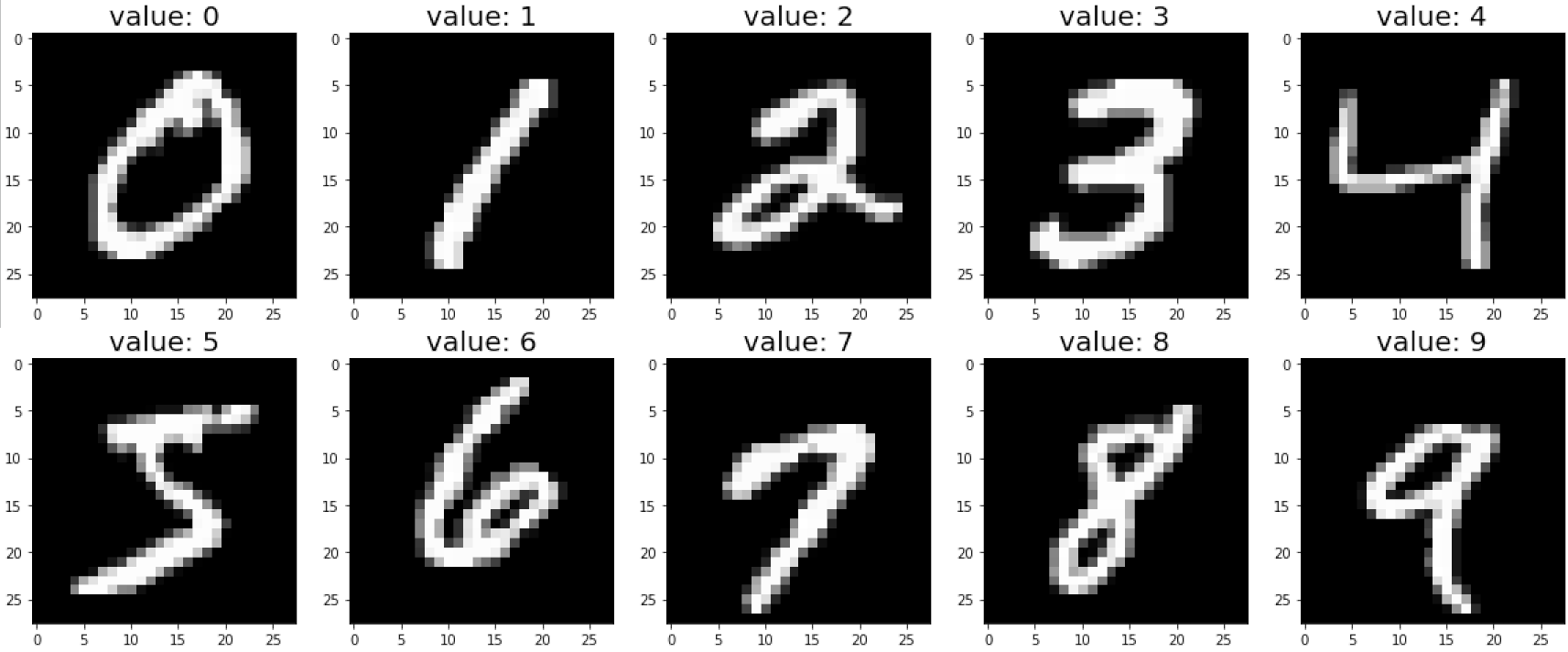}
    \label{fig:exemploMNIST}
\end{figure}

\subsubsection{What is the CIFAR-10 dataset}
\label{sec:funCifar10}

CIFAR-10, as MNIST, represents a large dataset of images; however, in the case of CIFAR-10, the image set is related to objects and animals. The present classes in the CIFAR-10 dataset are airplane, automobile, bird, cat, deer, dog, frog, horse, ship, and truck. It consists of 66,000 images, split 90\% for training and 10\% for testing. Each image comprises a 32x32x3 matrix, representing a 32 by 32 RGB image.

\begin{figure}[H]
    \centering
    \caption{Example of CIFAR-10 set images plotted graphically with the \textit{Matplotlib} tool}
    \includegraphics[width=\linewidth]{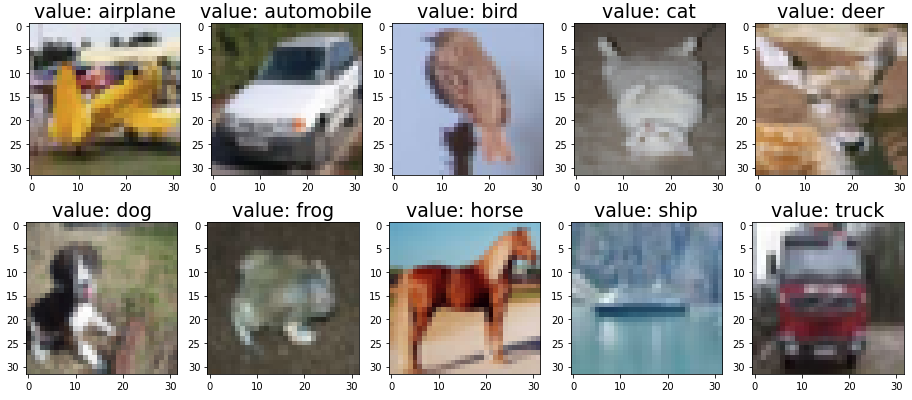}
    \label{fig:funExemploCIFAR}
\end{figure}

\subsection{Neural Networks}
\label{sec:funRedesNeurais}

It can be represented by a multi-layered directional graph, where each node is referred as a neuron and is responsible for an mathematical operation.

Artificial neural networks are systems inspired by biological neural networks constructed for data processing. It can be represented by a multi-layered directional graph, where each node is referred as a neuron and is responsible for an
mathematical operation \cite{o2015introduction}.
A neural network is designed to be trained on a specific
dataset, performing changes on its node operations with the objective of searching the neuron weight set that has the lowest error compared to the desired labeled result on the dataset \cite{o2015introduction}.

\subsubsection{Convolutional Neural Networks}

A Convolutional Neural Network is a specific type of neural network where each node is responsible for a convolution operation on the received input. An image vector is used as input and processed through the network into a final score represented on the output layer  \cite{o2015introduction}.

\subsection{Trained Model Structures}

The CNNs used to compare the performance metrics between Python and Elixir are present in the Axon library examples designed for model creation and training.

\subsubsection{MNIST Model Strucute}

This CNN is a straightforward network with two dense layers and a dropout between them\footnote{Available at: \url{https://github.com/elixir-nx/axon/blob/main/examples/vision/mnist.exs}}, as exemplified:

\begin{lstlisting}[language=elixir]
Axon.input({nil, 784}, "input")
  |> Axon.dense(128, activation: :relu)
  |> Axon.dropout(rate: 0.5)
  |> Axon.dense(10, activation: :softmax)
\end{lstlisting}

\subsubsection{CIFAR-10 Model Strucute}

It consists of a CNN on two convolutional layers followed by batch normalization and max-pooling layers and two dense layers with a dropout between them to avoid overfitting \footnote{Available at: \url{https://github.com/elixir-nx/axon/blob/main/examples/vision/cifar10.exs}}, as exemplified:

\begin{lstlisting}[language=elixir]
Axon.input({nil, 3, 32, 32}, "input")
  |> Axon.conv(32, kernel_size: {3, 3}, activation: :relu)
  |> Axon.batch_norm()
  |> Axon.max_pool(kernel_size: {2, 2})
  |> Axon.conv(64, kernel_size: {3, 3}, activation: :relu)
  |> Axon.batch_norm()
  |> Axon.max_pool(kernel_size: {2, 2})
  |> Axon.flatten()
  |> Axon.dense(64, activation: :relu)
  |> Axon.dropout(rate: 0.5)
  |> Axon.dense(10, activation: :softmax)
\end{lstlisting}

\section{Project}
\label{chap:projeto}

This section presents the three sub-projects that we created: 

\begin{itemize}
    \item An Elixir sub-project able to train CNNs able classify MNIST and CIFAR-10 datasets; 
    \item A Python sub-project able to train CNNs able classify MNIST and CIFAR-10 datasets; 
    \item  An Elixir sub-project able to extract performance benchmarks during the CNNs train. 
\end{itemize}

\begin{figure}[H]
    \centering
    \caption{High-level diagram exemplified the 3 projects interact.}
    \includegraphics[width=\linewidth]{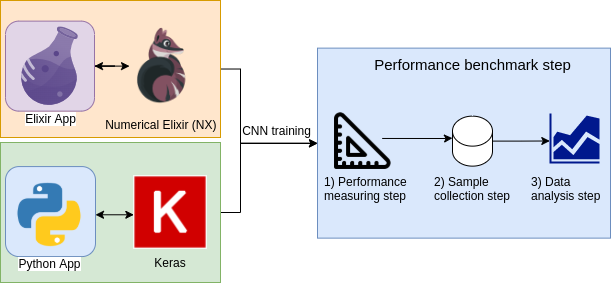}
    \label{fig:diagramaAltoNivel}
\end{figure}

Figure  \ref{fig:diagramaAltoNivel} is a visual representation of how the sub-projects interact and how they compose the needed stakeholders for this project. Also, we will explain the performance benchmark step in Section \ref{sec:experiment}.

\subsection{Elixir CNN trainer}
\label{subsec:elixirProject}

This sub-project consists of a code created with Mix\footnote{More information at: \url{https://elixirmix.com/}} using Elixir at version \textbf{1.12.3} and Erlang/OTP \textbf{24.0.6}. This sub-project source code is available on Github\footnote{Available at: \url{https://github.com/sallaumen/elixir_neural_network_labs}}. This sub-project mainly uses three libraries from the Nx\footnote{Available at: \url{https://github.com/elixir-nx}} ecosystem: Scidata, Axon, Exla.

\subsubsection{Scidata}

According to its documentation\footnote{Available at: \url{https://github.com/elixir-nx/scidata}}, this library is responsible for providing datasets in a simple way for Elixir projects. At this moment, it supports 11 different dataset types, of which two of them, MNIST and CIFAR-10, are used for this sub-project implementation.

\subsubsection{Axon}

According to its documentation\footnote{Available at: \url{https://github.com/elixir-nx/axon}}, this library is an Application Programming Interface (API) for Nx. At our project, Axon is used mainly for two functionalities:

\begin{itemize}
    \item Model Creation API: Axon provides a way to create a data structure for a CNN model;
    \item Training API: It provides a simple interface for training and evaluation.
\end{itemize}

\subsubsection{EXLA}

This library\footnote{Available at: \url{https://github.com/elixir-nx/nx/tree/main/exla}} is the Google’s Accelerated Linear Algebra (XLA) compiler for Nx, acting both as a backend for Nx tensors and as an \textbf{Nx.Defn}\footnote{\textbf{Nx.Defn} a new way to define methods in Elixir added to the language when using Nx} compiler \cite{elxa}. It is also the responsible library that allows Nx.Defn code to be compiled and executed at the GPU.

\subsection{Python CNN trainer}
\label{subsec:projPythonProject}

We developed the Python model\footnote{Available at: \url{https://github.com/sallaumen/python_neural_network_labs}} using Keras, a machine learning library built for model construction, training, and evaluation \cite{keras}. We built this sub-project as follows:

\begin{itemize}
    \item Dataset preparation: MNIST and CIFAR10 datasets are loaded similarly, using the dataset module from Keras, reshaped and normalized, so the RGB values are between 0 and 1.
    \item Model construction: Both models built for MNIST and CIFAR10 are built with the Sequential method from the models module, using convolutional and dense layers.
    \item Model compilation: We built the model using sparse categorical cross-entropy as a loss function, SGD as an optimizer, and accuracy as a metric.
    \item Model training: We trained the model with three epochs and batch size set as default to 32. 
\end{itemize}

As the purpose of this work is to analyze performance and qualitative metrics, we will not evaluate the accuracy of this algorithm.

\subsection{UnixStats: Performance benchmark data exporter}
\label{subsec:elixirBenchmark}
\begin{figure}[H]
    \centering
    \caption{Example of how UnixStats ":pretty" format is printed at the Linux terminal.}
    \includegraphics[width=\linewidth]{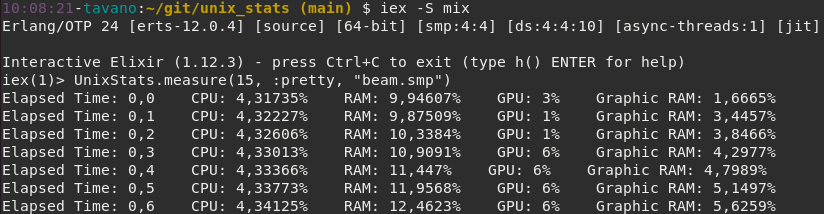}
    \label{fig:unix_stats_term}
\end{figure}

UnixStats is another Elixir sub-project developed for this study. It is a terminal-only software able to measure the use of CPU, RAM, GPU, and Graphic RAM from the executing machine. To run it, the user needs to provide three inputs:

\begin{itemize}
    \item  The "measured process" represents the name of the process that UnixStats will be measuring, which for Elixir was "beam.smp" and for Python, it was "python3.8". If the process is not running currently, UnixStats admits the Graphic RAM value as zero; 
    \item The "visualization type" can be ":pretty" or ":csv". The chosen format defines how the measured data will be presented to the software user. The image \ref{fig:unix_stats_term} exemplifies how it looks at the terminal; 
    \item The "monitoring time" represents how long UnixStats should keep measuring the machine.
\end{itemize}

\section{Experiment}
\label{sec:experiment}

This section describes how the experiment was prepared, executed and how data analysis was possible using the measured benchmarks. 

\subsection{Preparation}
\label{sec:expPrep}

The environment used to train the neural network in Elixir and Python is the same, a personal computer using Linux Ubuntu operational system, with no graphical interface, configured with the needed drivers, as described at 4.2, in a machine containing the following hardware:

\begin{itemize}
    \item Motherboard: ASUSTeK P8H61-M LX3 Plus R2.0
    \item CPU: Intel Core i5 3570 3.40GHz
    \item Graphics card: NVIDIA GeForce RTX 3060 Ti
    \item RAM: 16GB DDR3 1333MHz
    \item Power source potency: 850W
\end{itemize}

\subsection{GPU training environment preparation}
\label{subsec:gpuEnvPrep}

A fast way to train a CNN model is by using GPU and taking advantage of its processing power. Therefore, the used OS had to be prepared to run Elixir and Python CNN model training datasets. To be able to use the GPU, the following driver, Compute Unified Device Architecture (CUDA), and NVIDIA CUDA® Deep Neural Network library (cuDNN) were installed: 

\begin{itemize}
    \item NVIDIA video driver version: 470.103.01
    \item NVIDIA CUDA version: 11.4
    \item NVIDIA cuDNN version: 11.2
\end{itemize}

Finally the preparation of the local environment to run the projects made in Elixir and Python was completed. 

\subsection{Execution}
\begin{enumerate}
    \item We prepared the machine described above;
    \item Initialized the OS where only a terminal screen gets started;
    \item Using the software tmux, we divided the terminal in 2 independent sides;
    \item We initialized the software UnixStats at the left terminal;
    \item Then we used the right terminal to execute the CNN trainings in Elixir and Python with both analyzed datasets;
    \item We've trained CNN models twenty times, five for each combination of the studied languages: Elixir and Python, and datasets: MNIST and CIFAR-10, while \textit{UnixStats} software was used to measure each execution performance benchmark;
    \item For each run measured, UnixStats generated one file in Comma-separated Values (CSV) format;
    \item All CSVs then got imported Microsoft Excel software;
    \item Since every measured execution had a variable initial time due to the dataset download time, we had to do data normalization to minimize the download variation time effect at our measured data;
    \item For each group of five samples, we've created a sixth sample with the use average of CPU, RAM, GPU, and Video RAM;
    \item At Last, with the samples from the experiments done and the average data, we made graphs to enable the performance comparison aimed at this study, one for each combination group
\end{enumerate}

\section{Results}
\label{sec:resultAnaly}

In this section, we compared the CNN's training datasets using the Elixir and the Python languages. The metrics analyzed were the training time, and how much used of CPU, RAM, GPU, and Graphic RAM during the experiments.

\subsection{Evaluation}
\label{sec:evatualtion}

Below are the general performance graphs of training neural networks trained for datasets MNIST and CIFAR-10 in Elixir and Python. The comparison and analysis of these data will be further analyzed in this section.

General performance graphs of the two programming languages when training their neural networks to solve the MNIST dataset model.

\begin{figure}[H]
    \centering
    \caption{Graph of the system usage during model training MNIST in Elixir.}
    \includegraphics[width=\linewidth]{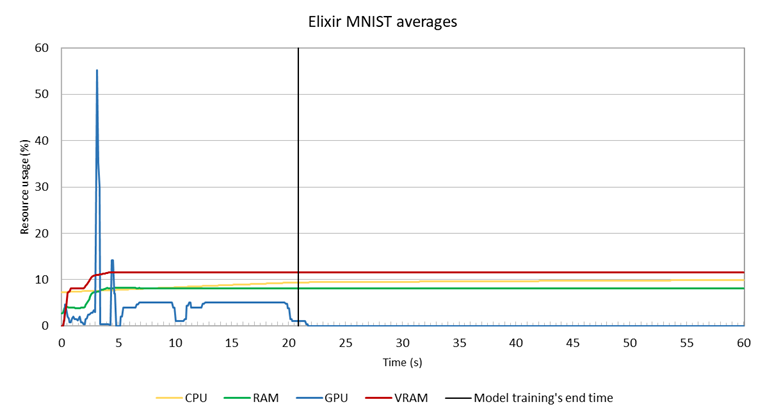}
    \label{fig:mnistAllDataElixir}
\end{figure}

\begin{figure}[H]
    \centering
    \caption{Graph of system usage during model training MNIST in Python.}
    \includegraphics[width=\linewidth]{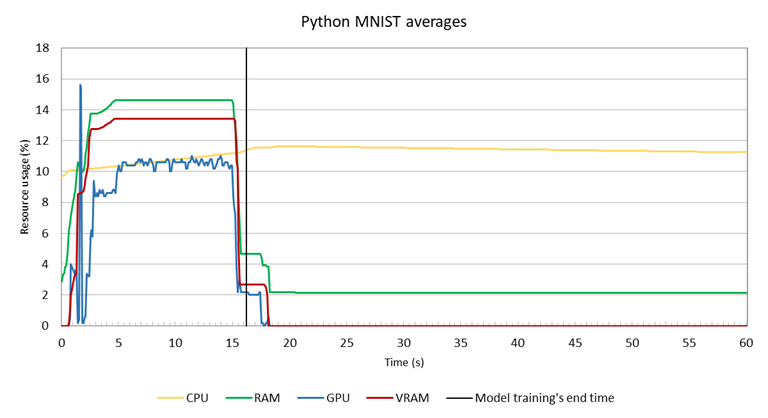}
    \label{fig:mnistAllDataPython}
\end{figure}

\begin{figure}[H]
    \centering
    \caption{Graph of system usage during model training CIFAR-10 in Elixir.}
    \includegraphics[width=\linewidth]{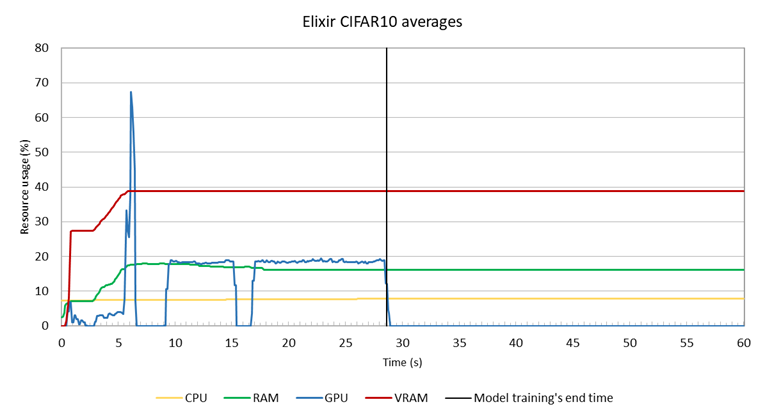}
    \label{fig:cifarAllDataElixir}
\end{figure}

\begin{figure}[H]
    \centering
    \caption{Graph of system usage during model training CIFAR-10 in Python.}
    \includegraphics[width=\linewidth]{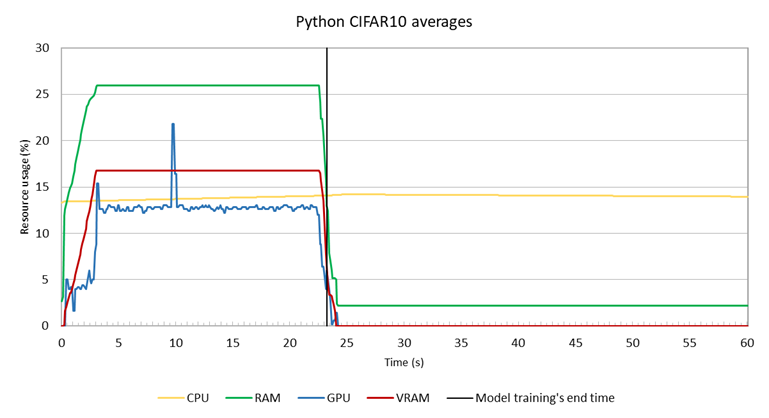}
    \label{fig:cifarAllDataPython}
\end{figure}

\subsection{Training time}

In this session, the overall training time will be analyzed in table \ref{tab:resTempo}, composed of three columns. The "Language" column refers to the programming language used, and the column "Dataset" contains the dataset used. Lastly, "Training time" has the total time used to train the neural network model capable of classifying images from the given dataset.

\begin{table}[H]
    \centering
    \caption{Table with training times of datasets MNIST and CIFAR-10 in Elixir and Python.}
    \begin{tabular}{|l|l|l|}
    \hline
    Language  & Dataset   & Training time (s) \\ \hline
    Elixir    & MNIST    & 20,4               \\ \hline
    Python    & MNIST    & 15,2               \\ \hline
    Elixir    & CIFAR-10 & 28,5               \\ \hline
    Python    & CIFAR-10 & 23,0               \\ \hline
    \end{tabular}
    \label{tab:resTempo}
\end{table}

Table \ref{tab:resTempo} shows that the MNIST dataset had 34.21\% greater time in Elixir than the same dataset trained in Python. While for CIFAR-10, the training time in Elixir was 23.91\% greater than in Python. 

\subsection{CPU usage}
\label{sec:resCpu}

\begin{figure}[H]
    \centering
    \caption{Graph comparing CPU usage while training using the MNIST dataset.}
    \includegraphics[width=\linewidth]{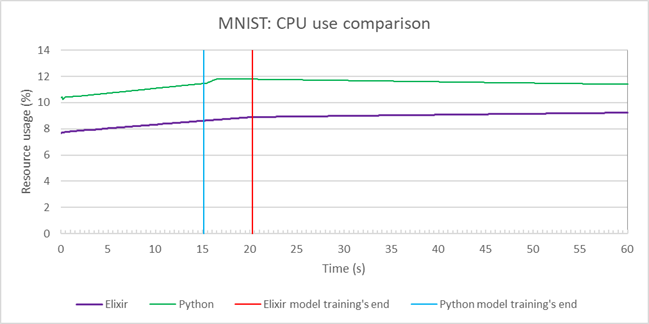}
    \label{fig:mnistCpu}
\end{figure}

\begin{figure}[H]
    \centering
    \caption{Graph comparing CPU usage while training using the CIFAR-10 dataset.}
    \includegraphics[width=\linewidth]{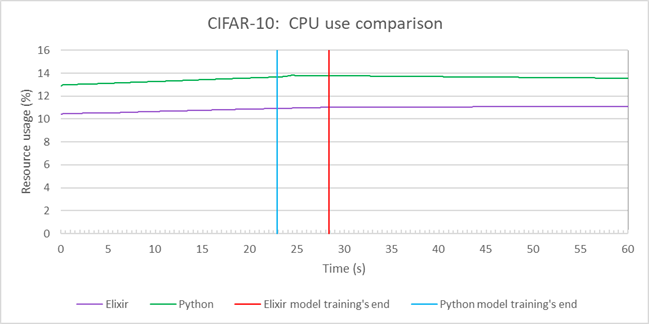}
    \label{fig:cifar10Cpu}
\end{figure}

In this session, we will present CPU usage while training the CNN, and for that, graphs \ref{fig:mnistCpu} and \ref{fig:cifar10Cpu} will be analyzed. While training in Elixir, CPU usage variation is 0.55\% when trained in the MNIST dataset and 0.36\% for CIFAR-10. While training in Python, CPU usage variation is 1.59\% when trained in the MNIST dataset and 0.91\% for CIFAR-10. It is possible to conclude, due to the low variance of this resource, that CPU seems not to be too relevant when training in both languages.

\subsection{RAM usage}
\label{sec:resRam}

\begin{figure}[H]
    \centering
    \caption{Graph comparing RAM usage while training using the MNIST dataset.}
    \includegraphics[width=\linewidth]{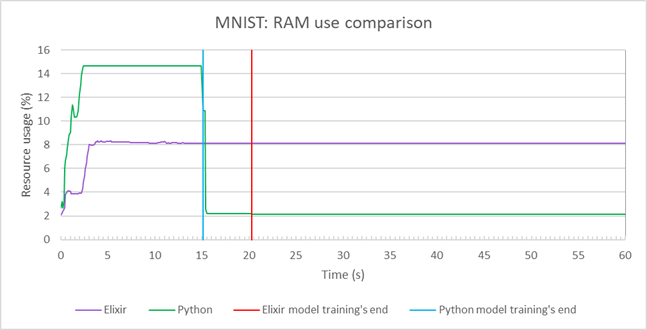}
    \label{fig:mnistRam}
\end{figure}

\begin{figure}[H]
    \centering
    \caption{Graph comparing RAM usage while training using the CIFAR-10 dataset.}
    \includegraphics[width=\linewidth]{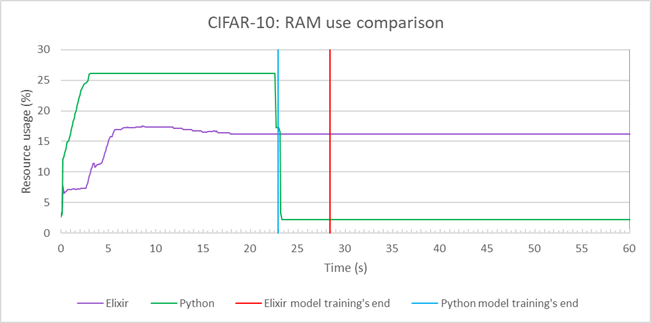}
    \label{fig:cifar10Ram}
\end{figure}

In this session, we will present RAM usage while training the CNN, and for that, graphs \ref{fig:mnistRam} and \ref{fig:cifar10Ram} will be analyzed. While training in Elixir, the variation of RAM usage is 7.2\% on MNIST and 15.9\% on CIFAR-10. While training in Python, the variation of RAM usage is 12.52\% on MNIST and 23.95\% on CIFAR-10.

It is possible to conclude that RAM is an essential resource during training in both languages, which was expected once the sets of images of both datasets are stored in memory during the execution. In addition, it is also possible to notice that in Elixir, there is less RAM usage than in Python, using an average of 42.4\% less of this resource for training on MNIST and 33.6\% less of this resource for CIFAR-10. Since the data size of datasets in both languages is the same, we can conclude that Elixir has better memory management in the analyzed scenarios.

\subsection{GPU usage}
\label{sec:resGpu}
\begin{figure}[H]
    \centering
    \caption{Graph comparing GPU usage while training using the MNIST dataset.}
    \includegraphics[width=\linewidth]{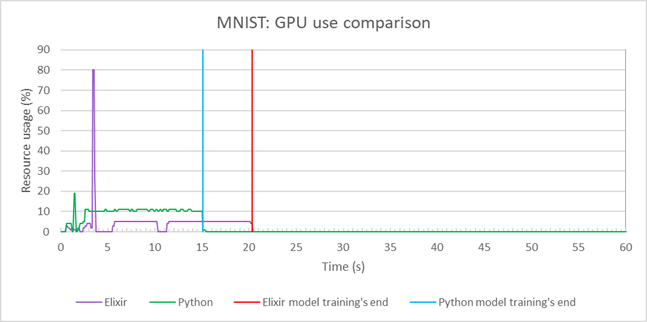}
    \label{fig:mnistGpu}
\end{figure}

\begin{figure}[H]
    \centering
    \caption{Graph comparing GPU usage while training using the CIFAR-10 dataset.}
    \includegraphics[width=\linewidth]{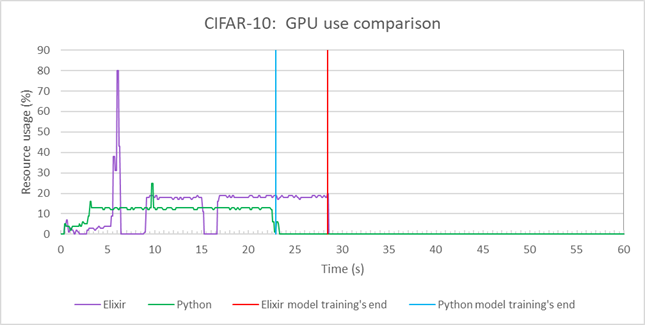}
    \label{fig:cifar10Gpu}
\end{figure}

In this session, we will analyze the GPU usage for training, and graphs \ref{fig:mnistGpu} and \ref{fig:cifar10Gpu} will be examined. While training in Elixir, the variation of GPU usage is 54\% on MNIST and 78\% for CIFAR-10. Analyzing each step, GPU usage peaks at 2.2 seconds, reaching 54\%, dropping to 0 for 1 second, then the usage goes up again, keeping up about 10\% from 4.1 seconds to 24.1 seconds. While training in Python, the variation of GPU usage is 19\% on MNIST and 25\% for CIFAR-10. Analyzing each step, GPU usage peaks at 1.4 seconds (VALIDATE), reaching 16\% GPU usage, dropping to 0 for 1 second, then the usage rises again, maintaining 10\% from 3.4 seconds to 18.1 seconds. 

With this, we can draw preliminary conclusions from this analysis. The first point is that both languages follow a pattern of using approximately 10\% of GPU during model training. Furthermore, we can also conclude that in this context that training the model in Python was faster than in Elixir by approximately 25\%, finishing 6 seconds faster.

\subsection{Video RAM usage}
\label{sec:resGRam}

\begin{figure}[H]
    \centering
    \caption{Graph comparing GPU RAM usage while training using the MNIST dataset.}
    \includegraphics[width=\linewidth]{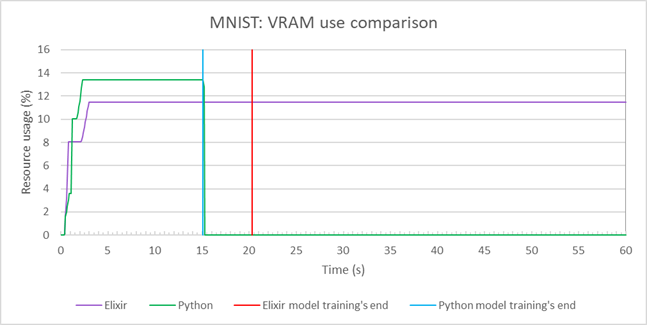}
    \label{fig:mnistGRam}
\end{figure}

\begin{figure}[H]
    \centering
    \caption{Graph comparing GPU usage while training using the CIFAR-10 dataset.}
    \includegraphics[width=\linewidth]{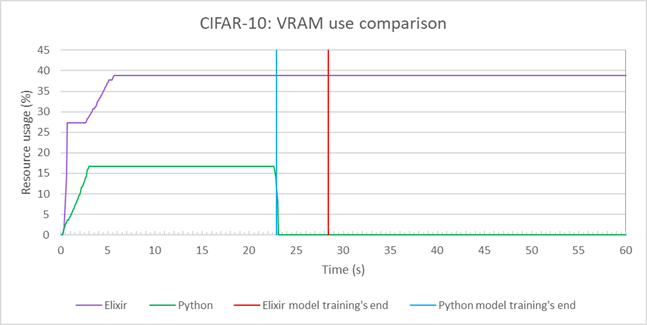}
    \label{fig:cifarGRam}
\end{figure}

In this section, we will analyze Video RAM use during training using graphs \ref{fig:mnistGRam} and \ref{fig:cifarGRam}. While training in Elixir, the variation in VRAM is 6.70\% for MNIST and 16.48\% for CIFAR-10. While training in Python, the variation in Video RAM is 13.42\% for MNIST and 16.78\% for CIFAR-10. By this initial analysis, it is possible to notice that for MNIST dataset, Elixir used less GPU memory, and for CIFAR-10 both languages used almost the same.

Also, it is possible to notice that both CNN trained in Elixir never deallocate the VRAM, this is caused by the compiler Accelerated Linear Algebra (XLA)\cite{50530}, which keeps this memory allocated until the process that it is associated ends, and since Elixir runs at the Erlang virtual machine: Bogdan/Björn’s Erlang Abstract Machine (BEAM), it only deallocates the VRAM when this virtual machine gets shutdown, which at our experiments we never done.

This behavior was also explained to us by the author of "Genetic Algorithms in Elixir", Sean Moriarity, when we made a post at Elixir Forum commenting about this behavior which until then we thought it was a bug\footnote{Elixir Forum post available at: \url{https://elixirforum.com/t/possible-graphic-ram-deallocation-issue-noticed-when-using-nx-with-exla/47629}}.

This issue does not happen with the Python trainer, since it runs as a script instead of in a virtual machine, making XLA deallocate the graphic RAM almost instantly after finishing the training, noticeable at the graphs \ref{fig:mnistGRam} and \ref{fig:cifarGRam} at the "Python model training's end".

\section{Conclusion}
\label{sec:conclusao}

In this article, we concluded that Elixir took more time in both datasets analyzed compared to Python while demanding more Graphic RAM on peaks. We observed that RAM usage is an exception to the other metrics as Elixir consumed on average 42\% and 33\% less RAM than Python, showing that Elixir has better memory management for these test cases.

As for usability, we observed that Python has much more documentation, material, and available libraries for dataset preparation, model creation, and training. This was expected as Python is already much more consolidated in the machine learning market. However, the Nx library was a significant improvement to Elixir, allowing the language to be used for neural network creations much more efficiently than before Nx.


\section{Acknowledgments}
We are grateful to José Valim, Sean Moriarity, and every other developer working on the Numerical Elixir ecosystem. It has been incredible for the Elixir language, and it is incredible to see how Nx has been improving since its announcement in February 2021.

\bibliographystyle{plain} 
\bibliography{main} 

\begin{thebibliography}{1}

\bibitem{allamanis2018survey}
Miltiadis Allamanis, Earl~T Barr, Premkumar Devanbu, and Charles Sutton.
\newblock A survey of machine learning for big code and naturalness.
\newblock {\em ACM Computing Surveys (CSUR)}, 51(4):1--37, 2018.

\bibitem{IntroDatasets}
Google Cloud.
\newblock Introduction to datasets, 2022.

\bibitem{elxa}
EXLA.
\newblock Exla hexdocs, 2022.

\bibitem{keras}
Keras.
\newblock Keras, 2022.

\bibitem{o2015introduction}
Keiron O'Shea and Ryan Nash.
\newblock An introduction to convolutional neural networks.
\newblock {\em arXiv preprint arXiv:1511.08458}, 2015.

\bibitem{50530}
Amit Sabne.
\newblock Xla : Compiling machine learning for peak performance, 2020.

\end{thebibliography}

\end{document}